\documentclass{article}

\usepackage{url}            
\usepackage{booktabs}       
\usepackage{amsfonts}       
\usepackage{nicefrac}       
\usepackage{microtype}      

\usepackage{amsmath,amssymb,amsthm,amsfonts}
\usepackage{algorithm,algorithmic}
\usepackage{changepage}
\usepackage{graphicx}

\newtheorem{thm}{Theorem}

\newtheorem{cor}{Corollary}

\newcommand{\bE}{\mathbb{E}}
\newcommand{\bP}{\mathbb{P}}
\newcommand{\bR}{\mathbb{R}}

\newcommand{\sS}{{\mathcal S}}
 
\newcommand{\sX}{{\mathcal X}} 
 
\newcommand{\sZ}{{\mathcal Z}} 
 
\newcommand{\sN}{{\mathcal N}}

\newcommand{\argmin}{\operatornamewithlimits{arg\ min}}
\newcommand{\argmax}{\operatornamewithlimits{arg\ max}}

\newcommand{\set}[1]{\left\{#1\right\}}

\newcommand{\norm}[1]{\left\|#1\right\|}

\def\ie{{\em i.e.,~}}
\def\eg{{\em e.g.,~}}

\usepackage[usenames]{color}

\newcommand{\algname}[1]{\textnormal{\textsc{#1}}}

\usepackage{times}
\usepackage{graphicx} 
\usepackage{subfigure} 

\usepackage{natbib}

\usepackage{algorithm}
\usepackage{algorithmic}

\usepackage{hyperref}

\usepackage[accepted]{icml2017}

\icmltitlerunning{Leveraging Union of Subspace Structure to Improve Constrained Clustering}

\begin{document} 

\twocolumn[
    \icmltitle{Leveraging Union of Subspace Structure to Improve Constrained Clustering}



\icmlsetsymbol{equal}{*}

\begin{icmlauthorlist}
\icmlauthor{John Lipor}{um}
\icmlauthor{Laura Balzano}{um}
\end{icmlauthorlist}

\icmlaffiliation{um}{Department of Electrical and Computer Engineering, University Michigan, Ann Arbor, MI, USA}

\icmlcorrespondingauthor{John Lipor}{lipor@umich.edu}

\icmlkeywords{active learning, constrained clustering, subspace clustering, union of subspaces}

\vskip 0.3in
]



\printAffiliationsAndNotice{}  

\begin{abstract} 
Many clustering problems in computer vision and other contexts are also classification problems, where each cluster shares a meaningful label. Subspace clustering algorithms in particular are often applied to problems that fit this description, for example with face images or handwritten digits. While it is straightforward to request human input on these datasets, our goal is to reduce this input as much as possible.
We present a pairwise-constrained clustering algorithm that actively selects queries based on the union-of-subspaces model.
    The central step of the algorithm is in querying \emph{points of minimum margin} between estimated subspaces; analogous to classifier margin, these lie near the decision boundary. We prove that points lying near the intersection of subspaces are points with low margin. Our procedure can be used after any subspace clustering algorithm that outputs an affinity matrix. We demonstrate on several datasets that our algorithm drives the clustering error down considerably faster than the state-of-the-art active query algorithms on datasets with subspace structure and is competitive on other datasets. 
\end{abstract} 

\section{Introduction}

The union of subspaces (UoS) model, in which data vectors lie near one of several subspaces, has been used actively in the computer vision community on datasets ranging from images of objects under various lighting conditions \cite{basri2003lambertian} to visual surveillance tasks \cite{oliver2000bayesian}. The recent textbook \cite{vidal2016generalized} includes a number of useful applications for this model, including lossy image compression, clustering of face images under
different lighting conditions, and video segmentation. Subspace clustering algorithms utilize the UoS model to cluster data vectors and estimate the underlying subspaces, achieving excellent performance on a variety of real datasets.
However, as we will show in Section \ref{sec:simulations}, even oracle UoS
classifiers do not achieve perfect clustering on these datasets. While current algorithms for subspace clustering are unsupervised, in many cases a human could provide relevant information in the form of pairwise constraints between points, \eg answering whether two images are of the same person or whether two objects are the same.

The incorporation of pairwise constraints into clustering algorithms is known as pairwise-constrained clustering (PCC). PCC algorithms use supervision in the form of \textit{must-link} and \textit{cannot-link} constraints by ensuring that points with must-link constraints are clustered together and points with cannot-link constraints are clustered apart. In \cite{davidson2006measuring}, the authors investigate the phenomenon that incorporating poorly-chosen
constraints can lead to an increase in clustering error, rather than a decrease as one would expect from additional label information. This is because points constrained to be in the same cluster that are otherwise dissimilar can confound the constrained clustering algorithm. For this reason, researchers have turned to \textit{active} query selection methods, in which constraints are intelligently selected based on a number of heuristics. 
These algorithms perform well across a number of datasets but do not take advantage of any known structure in the data. In the case where data lie on a union of subspaces, one would hope that knowledge of the underlying geometry could give hints as to which points are likely to be clustered incorrectly.

\begin{figure}[t]
    \centering
    \includegraphics[width=0.6\columnwidth]{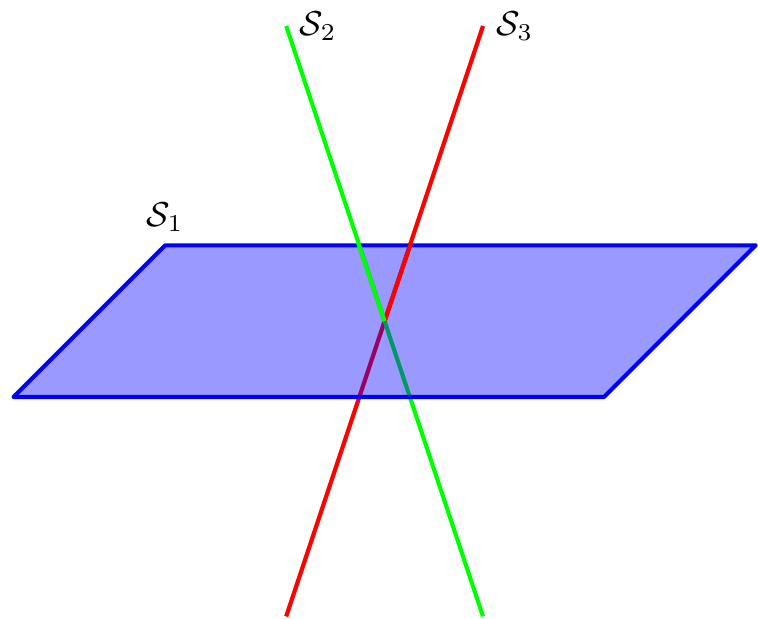}
    \caption{Example union of $K = 3$ subspaces of dimensions $d_{1} = 2$, $d_{2} = 1$, and $d_{3} = 1$.}
    \label{fig:uos}
    \vspace{-1em}
\end{figure}

Let $\sX = \set{x_{i} \in \bR^{D}}_{i=1}^{N}$ be a set of data points lying near a union of $K$ linear subspaces of the ambient space. We denote the subspaces by $\set{\sS_{k}}_{k=1}^{K}$, each having dimension $d_{k}$. An example union of subspaces is shown in Fig. \ref{fig:uos}, where $d_{1} = 2$, $d_{2} = d_{3} = 1$. The goal of subspace clustering algorithms has traditionally been to cluster the points in $\sX$ according to their nearest subspace without any supervised input. We turn this around and ask whether this model is useful for active clustering, where we request a very small number of intelligently selected labels. A key observation when considering data well-modeled by a union of subspaces is that uncertain points will be ones lying equally distant to multiple subspaces. Using a novel definition of margin tailored for the union of subspaces model, we incorporate this observation into an active subspace clustering algorithm. 

Our contributions are as follows. We introduce a novel algorithm for pairwise constrained clustering that leverages UoS structure in the data. A key step in our algorithm is choosing points of \textit{minimum margin}, \ie those lying near a decision boundary between subspaces. We define a notion of margin for the UoS model and provide theoretical insight as to why points of minimum margin are likely to be misclustered by unsupervised algorithms. We show through extensive experimental results that when the data lie
near a union of subspaces, our method drastically
outperforms existing PCC algorithms, requiring far fewer queries to achieve perfect clustering. Our datasets range in dimension from 256-2016, number of data points from 320-9298, and number of subspaces from 5-100. On ten MNIST digits with a modest number of queries, we get 5\% classification error with only 500 pairwise queries compared to about 20\% error for current state-of-the-art PCC algorithms and 35\% for unsupervised algorithms. We also achieve 0\% classification error on the full Yale, COIL, and USPS datasets with a small fraction of the number of queries needed by competing algorithms.  In datasets where we do not expect subspace structure, our algorithm still achieves competitive performance.
Further, our algorithm is agnostic to the input subspace clustering algorithm and can therefore take advantage of any future algorithmic advances for subspace clustering.

\section{Related Work}

A survey of recently developed subspace clustering algorithms can be found in \cite{vidal2011subspace} and the textbook \cite{vidal2016generalized}. In these and more recent work, clustering algorithms that employ spectral methods achieve the best performance on most datasets. Notable examples of such algorithms include Sparse Subspace Clustering (SSC) \cite{elhamifar2013sparse} and its extensions \cite{you2016scalable,you2016oracle}, Low-Rank Representation (LRR) \cite{liu2010robust}, Thresholded Subspace Clustering (TSC)
\cite{heckel2015robust}, and Greedy Subspace Clustering (GSC) \cite{park2014greedy}. Many recent algorithms exist with both strong theoretical guarantees and empirical performance, and a full review of all approaches is beyond the scope of this work. However, the core element of all recent algorithms lies in the formation of the affinity matrix, after which spectral clustering is performed to obtain label estimates. In SSC, the affinity matrix is formed via a series of $\ell_{1}$-penalized regressions. LRR uses a similar cost function but penalizes the nuclear norm instead of the $\ell_{1}$. TSC thresholds the spherical distance between points, and GSC works by successively (greedily) building subspaces from points
likely to lie in the same subspace. Of these methods, variants of SSC achieve the best overall performance on benchmark datasets and has the strongest theoretical guarantees, which were introduced in \cite{elhamifar2013sparse} and
strengthened in numerous recent works \cite{soltanolkotabi2012geometric,soltanolkotabi2014robust,wang2013noisy,wang2016graph}. 
While the development of efficient algorithms with stronger guarantees has received a great deal of attention, very little attention has been paid to the question of what to do about data that cannot be correctly clustered.
Thus, when reducing clustering error to zero (or near zero) is a priority, users must look beyond unsupervised subspace clustering algorithms to alternative methods. One such method is to request some supervised input in the form of pairwise constraints, leading to the study of pairwise-constrained clustering (PCC).

PCC algorithms work by incorporating \textit{must-link} and \textit{cannot-link} constraints between points, where points with must-link constraints are forced (or encouraged in the case of spectral clustering) to be clustered together, and points with cannot-link constraints are forced to be in separate clusters. In many cases, these constraints can be provided by a human labeler. For example, in \cite{biswas2014active}, the authors perform experiments where comparisons between human faces are provided
by users of Amazon Mechanical Turk with an error rate of 1.2\%. Similarly, for subspace clustering datasets such as Yale B and MNIST, a human could easily answer questions such as, ``Are these two faces the same person?'' and ``Are these two images the same number?'' An early example of PCC is found in \cite{wagstaff2001constrained}, where the authors modify the $K$-means cost function to incorporate such constraints. In \cite{basu2004active}, the authors utilize active
methods to initialize $K$-means in an intelligent ``\algname{Explore}" phase, during which neighborhoods of must-linked points are built up. After this phase, new points are queried against representatives from each neighborhood until a must-link is obtained. A similar explore phase is used in \cite{mallapragada2008active}, after which a min-max approach is used to select the most uncertain sample. Early work on constrained spectral clustering appears in
\cite{xu2005active,wang2010active}, in which spectral clustering is improved by examining the eigenvectors of the affinity matrix in order to determine the most informative points. However, these methods are limited to the case of two clusters and therefore impractical in many cases. 

More recently, the authors in \cite{xiong2016active,biswas2014active} improve constrained clustering by modeling which points will be most informative given the current clustering, with state-of-the-art
results achieved on numerous datasets by the algorithm in \cite{xiong2016active}, referred to as Uncertainty Reducing Active Spectral Clustering (URASC). URASC works by maintaining a set of \textit{certain sets},
whereby points in the same certain set are must-linked and points in different certain sets are cannot-linked. A test point $x_{T}$ is selected via an uncertainty-reduction model motivated by matrix perturbation theory, after which queries
are presented in an intelligent manner until $x_{T}$ is either matched with an existing certain set or placed in its own new certain set. In practice \cite{xiong2016personal}, the certain sets are initialized using the \algname{Explore} algorithm of \cite{basu2004active}. 


While we are certainly not the first to consider actively selecting labels to improve clustering performance, to the best of our knowledge we are the first to do so with structured clusters. Structure within and between data clusters is often leveraged for unsupervised clustering \cite{wright2009robust}, and that structure is also leveraged for adaptive sampling of the structured signals themselves (\eg see previous work on sparse \cite{haupt2011distilled, indyk2011power}, structured sparse \cite{soni2014fundamental}, and low rank signals \cite{krishnamurthy2013low}). This paper emphasizes the power of that structure for reducing the number of required labels in an active learning algorithm as opposed to reducing the number of samples of the signal itself, and points to exciting open questions regarding the tradeoff between signal measurements and query requirements in semi-supervised clustering.

\section{UoS-Based Pairwise-Constrained Clustering}

\begin{figure*}[t]
    \centering
    \includegraphics[width=0.7\linewidth]{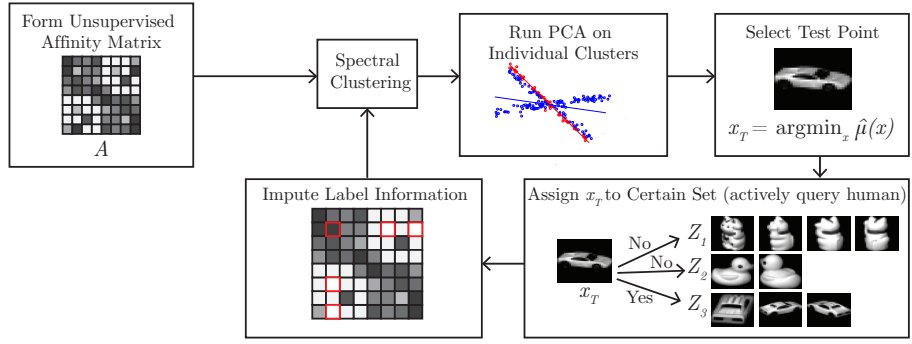}
    \caption{Diagram of SUPERPAC algorithm for pairwise constrained clustering.}
    \label{fig:algorithm}
    \vspace{-1em}
\end{figure*}

Recall that $\sX = \set{x_{i} \in \bR^{D}}_{i=1}^{N}$ is a set of data points lying on a union of $K$ subspaces $\set{\sS_{k}}_{k=1}^{K}$, each having dimension $d$. In this work, we assume all subspaces have the same dimension, but it is possible to extend our algorithm to deal with non-uniform dimensions. The goal is to cluster the data points according to this generative model, \ie assigning each data point to its (unknown) subspace. In this section we describe our algorithm, which actively selects pairwise constraints in order to improve clustering accuracy. The key step is choosing an informative query test point, which we do using a novel notion of \emph{minimum subspace margin}.

Denote the true clustering of a point $x \in \sX$ by $C(x)$. Let the output of a clustering algorithm (such as SSC) be an affinity/similarity matrix $A$ and a set of label estimates
$\set{\hat{C}(x_{i})}_{i=1}^{N}$. These are the inputs to our algorithm.
The high-level operation of our algorithm is as follows. 
To initialize, we build a set of certain sets $\sZ$ using an \algname{Explore}-like algorithm similar to that of \cite{basu2004active}. Certain sets are in some sense equivalent to labels in that points within a certain set belong to the same cluster and points across certain sets belong to different clusters. Following this, the following steps are repeated until a maximum number of queries has been made:
        \vspace{-1em}
\begin{enumerate}
    \item \textbf{Spectral Clustering:} Obtain label estimates via spectral clustering.
        \vspace{-0.5em}
    \item \textbf{PCA on each cluster:} Obtain a low-dimensional subspace estimate from points currently sharing the same estimated cluster label.
        \vspace{-0.5em}       
    \item \textbf{Select Test Point:} Obtain a test point $x_{T}$ using subspace margin with respect to the just estimated subspaces.        \vspace{-0.5em}
    \item \textbf{Assign $x_{T}$ to Certain Set:} Query the human to compare the test point with representatives from certain sets until a must-link is found or all certain sets have been queried, in which case the test point becomes its own certain set.
        \vspace{-0.5em}
    \item \textbf{Impute Label Information:} Certain sets are used to impute must-link and cannot-link values in the affinity matrix.
\end{enumerate}
        \vspace{-1em}
We refer to our algorithm as SUPERPAC (SUbsPace clustERing with Pairwise Active Constraints). A diagram of the algorithm is given in Fig. \ref{fig:algorithm}, and we outline each of these steps below and provide pseudocode in Algorithm \ref{alg:superpac}. 

\subsection{Sample Selection via Margin}

Min-margin points have been studied extensively in active learning; intuitively, these are points that lie near the decision boundary of the current classifier. In \cite{settles2012active}, the author notes that actively querying points of minimum margin (as opposed to maximum entropy or minimum confidence) is an appropriate choice for reducing classification error. 
In \cite{wang2016noise}, the authors present a margin-based binary classification algorithm that achieves an optimal
rate of convergence (within a logarithmic factor). 

\begin{figure}[t]
    \centering
    \includegraphics[width=2in]{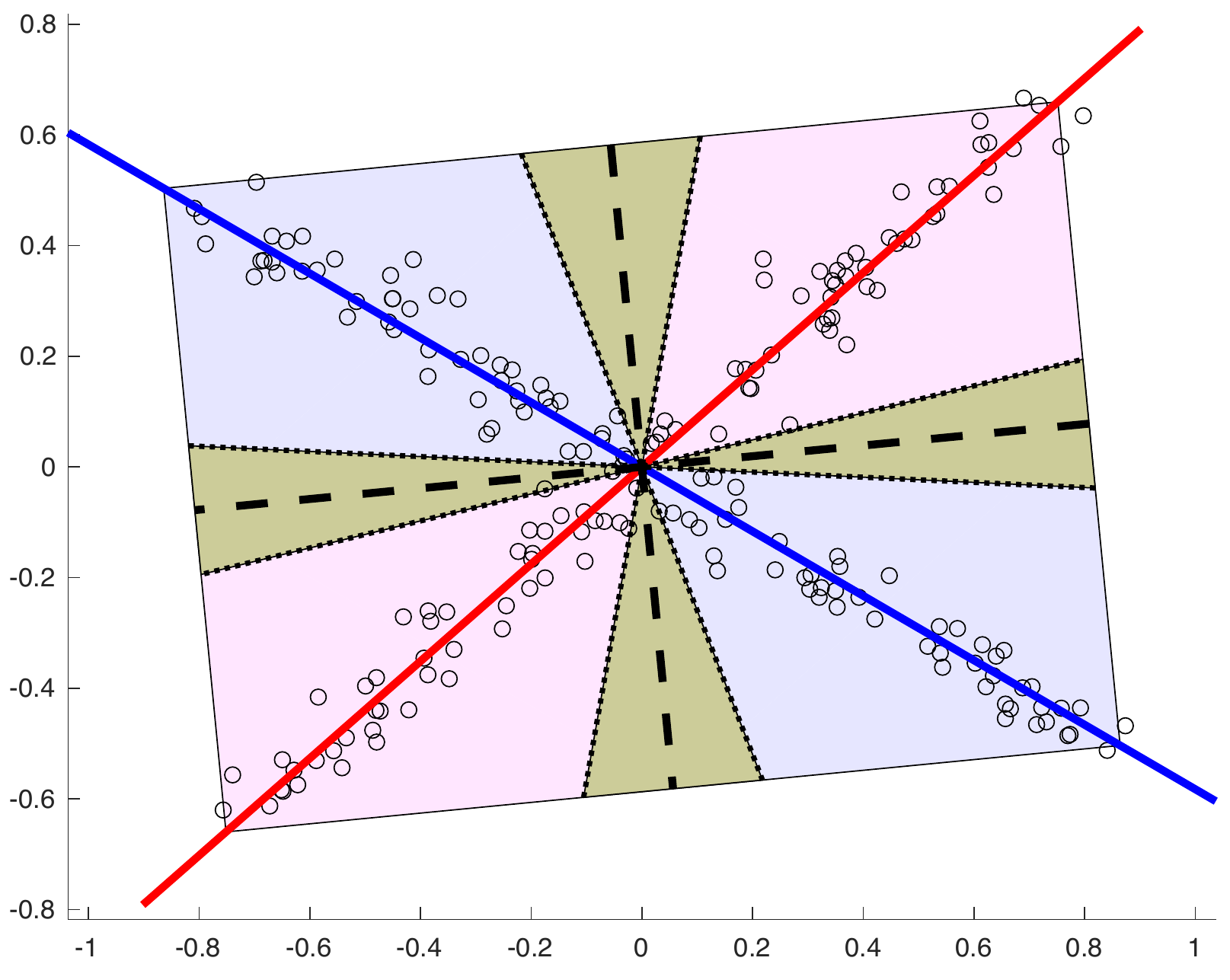}
    \vspace{-2mm}
    \caption{Illustration of subspace margin. The blue and red lines are the generative subspaces, with corresponding disjoint decision regions. The yellow-green color shows the region within some margin of the decision boundary, given by the dotted lines.}
    \label{fig:margin}
\end{figure}

In this section, we define a novel notion of margin for the UoS model and provide theoretical insight as to why points of minimum margin are likely to be misclustered. For a subspace $\sS_{k}$ with orthonormal basis $U_{k}$, let the distance of a point to that subspace be
    $\mathrm{dist}(x,\sS_{k}) = \min_{y \in \sS_k} \|x - y \|_2 = \norm{x - U_{k}U_{k}^{T}x}_{2}.$
Let $k^{*} = \argmin_{k \in [K]} \mathrm{dist}(x,\sS_{k})$ be the index of the closest subspace, where $[K] = \set{1,2,\cdots,K}$. Then the subspace margin of a point $x \in \sX$ is the ratio of closest and second closest subspaces, defined as
\begin{equation}
    \label{eq:ss_margin}
    \hat{\mu}(x) = 1 - \max_{j \neq k^{*}, j \in [K]} \frac{\mathrm{dist}(x,S_{k^{*}})}{\mathrm{dist}(x,S_{j})}.
\end{equation}
The point of minimum margin is then defined as $\argmin_{x \in \sX} \hat{\mu}(x)$.
Note that the fraction is a value in $[0,1]$, where the a value of 0 implies that the point $x$ is equidistant to its two closest subspaces. 
This notion is illustrated in Figure~\ref{fig:margin}, where the yellow-green color shows the region within some margin of the decision boundary.

In the following theorem, we show that points lying near the intersection of subspaces are included among those of minimum margin with high probability. This method of point selection is then motivated by the fact that the difficult points to cluster are those lying near the intersection of subspaces [12]. Further, theory for SSC ([11],[15]) shows that problematic points are those having large inner product with some or all directions in other subspaces. Subspace margin captures exactly this phenomenon.

\begin{thm}
    Consider two $d$-dimensional subspaces $\sS_{1}$ and $\sS_{2}$. Let $y = x + n$, where $x \in \sS_{1}$ and $n \sim \sN(0,\sigma^{2}I_{D})$. Define $$\mu(y) = 1 - \frac{\mathrm{dist}(y, \sS_1)}{\mathrm{dist}(y,\sS_2)}\;.$$ Then
    \begin{equation*}
        1 - \frac{(1+\varepsilon)\sqrt{\sigma^{2}(D-d)}}{(1-\varepsilon)\sqrt{\sigma^{2}(D-d) + \mathrm{dist}(x,\sS_2)^{2}}} \leq \mu(y)
    \end{equation*}
    and
    \begin{equation*}
        \mu(y) \leq 1 - \frac{(1-\varepsilon)\sqrt{\sigma^{2}(D-d)}}{(1+\varepsilon) \sqrt{\sigma^{2}(D-d) + \mathrm{dist}(x,\sS_2)^{2}}},
    \end{equation*}
    with probability at least $1 - 4e^{-c\varepsilon^{2}(D-d)}$, where $c$ is an absolute constant.
    \label{thm:jl} 
\end{thm}
The proof is given in Appendix \ref{app:thm}. Note that if $\mathrm{dist}(y, \sS_1) \leq \mathrm{dist}(y,\sS_2)$, then $\mu(y) = \hat{\mu}(y)$. In this case, Thm. \ref{thm:jl} states that under the given noise model, points with small residual to the incorrect subspace (\ie points near the intersection of subspaces) will have small margin. These are exactly the points for which supervised label information will be most beneficial.

The statement of Thm. \ref{thm:jl} allows us to quantify exactly how near a point must be to the intersection of two subspaces to be considered a point of minimum margin. Let $\phi_1 \leq \phi_2 \leq \cdots \leq \phi_d$ be the $d$ principal angles\footnote{See \cite{golub2012matrix} for a definition of principal angles.} between $\sS_1$ and $\sS_2$. If the subspaces are very far apart, $\frac{1}{d} \sum_{i=1}^d \sin^2(\phi_i)$ is near 1, and if they are very close $\frac{1}{d} \sum_{i=1}^d \sin^2(\phi_i)$ is near zero. Note that, for any $x \in \sS_1$,  $$\sin^2(\phi_1) \leq \mathrm{dist}(x,\sS_2)^2 \leq \sin^2(\phi_d)\;;$$ that is, there are bounds on $\mathrm{dist}(x,\sS_2)$ depending on the relationship of the two subspaces. 
We also know that if $x$ is drawn using isotropic Gaussian weights from a basis for $\sS_1$, 
then $$\bE \left[ \mathrm{dist}(x,\sS_2)^2 \right] = \frac{1}{d} \sum_{i=1}^d \sin^2(\phi_i)\;.$$ Given this, we might imagine that margin of the noisy points is a useful indicator of points near the intersection in a scenario where $\sin^2(\phi_1)$ is small but $\frac{1}{d} \sum_{i=1}^d \sin^2(\phi_i)$ is not, \eg when the subspaces have an intersection but are distant in other directions. With this in mind we state the following corollary, whose proof can be found in Appendix \ref{app:cor}.

\begin{cor} Suppose $x_1 \in \sS_1$ is such that 
    \begin{equation}
        \label{eq:delta}
        \mathrm{dist}(x_1,\sS_2)^2 = \sin^2(\phi_1) + \delta  \left(\frac{1}{d} \sum_{i=1}^d \sin^2(\phi_i) \right)
    \end{equation}
    for some small $\delta\geq 0$; that is, $x_1$ is close to the intersection of $\sS_1$ and $\sS_2$. Let $x_2$ be a random point in $\sS_1$ generated as $x_2 = U_1 w$ where $U_1$ is a basis for $\sS_1$ and $w \sim \sN(0,\frac{1}{d}I_{d})$. We observe $y_i = x_i + n_i$, where $n_i \sim \sN(0, \sigma^2 )$, $i=1,2$. 
    If there exists $\tau>1$ such that $$\delta < \frac{5}{7} - \frac{1}{\tau}$$ and 
    \begin{equation}
        \label{eq:tau}
        \tau \left(\sin^2(\phi_1) + \frac{1}{6}\sigma^2\left(D-d\right)\right) < \frac{1}{d} \sum_{i=1}^d \sin^2(\phi_i)\;,
    \end{equation}
    that is, the average angle is sufficiently larger than the smallest angle, then
    $$\bP\set{\mu(y_1) < \mu(y_2)} \geq 1-e^{-c \left(\frac{7}{100}\right)^2 d s} - 4e^{-c\left(\frac{1}{50}\right)^{2}(D-d)}$$ where $\mu(y)$ is defined as in Thm. \ref{thm:jl}, $c$ is an absolute constant, and $s =  \frac{1}{d} \sum_{i=1}^d \sin^2(\phi_i)$.
    \label{cor:mu}
\end{cor}

We make some remarks first to connect our results to other subspace distances that are often used. Perhaps the most intuitive form of subspace distance between that spanned by $U_1$ and $U_2$ is $\frac{1}{d} \|(I-U_1U_1)^TU_2\|_F^2$; if the two subspaces are the same, the projection onto the orthogonal complement is zero; if they are orthogonal, we get the norm of $U_2$ alone, giving a distance of 1. This is equal to the more visually symmetric $1-\frac{1}{d} \|U_1^TU_2\|_F^2$, another common
distance. Further we note that, by the definition of principal angles \cite{golub2012matrix}, $$1-\frac{1}{d} \|U_1^TU_2\|_F^2 = 1-\frac{1}{d} \sum_{i=1}^d \cos^2(\phi_i) = \frac{1}{d} \sum_{i=1}^d \sin^2(\phi_i)\;.$$

From Equation \eqref{eq:delta}, we see that the size of $\delta$ determines how close $x_1 \in \sS_1$ is to $\sS_2$; if $\delta=0$, $x_1$ is as close to $\sS_2$ as possible. For example, if $\phi_1 = 0$, the two subspaces intersect, and $\delta=0$ implies that $x_1 \in \sS_1 \cap \sS_2$. Equation \eqref{eq:tau} captures the gap between average principal angle and the smallest principal angle. We conclude that if this gap is large enough and $\delta$ is small enough so that $x_1$ is close to $\sS_2$, then the observed $y_1$ will have smaller margin than the average point in $\sS_1$, even when observed with noise. 


For another perspective, consider that in the noiseless case, for $x_1,x_2 \in \sS_1$, the condition $\mathrm{dist}(x_1,\sS_2) < \mathrm{dist}(x_2,\sS_2)$ is enough to guarantee that $x_{1}$ lies nearer to $\sS_2$. 
Under the given additive noise model ($y_{i} = x_{i} + n_{i}$ for $i = 1,2$) the gap between $\mathrm{dist}(x_1,\sS_2)$ and $\mathrm{dist}(x_2,\sS_2)$ must be larger by some factor depending on the noise level. After two applications of Thm. \ref{thm:jl} and rearranging terms, we have that $\mu(y_{1}) < \mu(y_{2})$ with high probability if 
\begin{equation}
    \beta \mathrm{dist}(x_2,\sS_2)^{2} - \mathrm{dist}(x_1,\sS_2)^{2} > (1-\beta) \sigma^{2}(D-d).
    \label{eq:gap}
\end{equation}
where $\beta = \left((1-\varepsilon)/(1+\varepsilon)\right)^4$, a value near 1 for small $\varepsilon$. Equation \eqref{eq:gap} shows that the gap $\mathrm{dist}(x_2,\sS_2)^{2} - \mathrm{dist}(x_1,\sS_2)^{2}$ must grow (approximately linearly) with the noise level $\sigma^{2}$. The relationship of this gap to the subspace distances is quantified by Corollary \ref{cor:mu}; plugging $\sin^{2}(\phi_{1})$ from Equation \eqref{eq:delta} into Equation \eqref{eq:tau} and rearranging yields a statement of the form in Equation \eqref{eq:gap}. 




\begin{algorithm}[t]
    \caption{\algname{SUPERPAC}}
    \label{alg:superpac}
    \begin{algorithmic}
        \STATE \textbf{Input:} $\sX = \set{x_{1},x_{2},\dots,x_{N}}$: data, $K$: number of clusters, $d$: subspace dimension, $A$: affinity matrix, maxQueries: maximum number of pairwise comparisons
        \STATE \textbf{Estimate Labels:} $\hat{C} \gets$ \algname{SpectralClustering}($A$,$K$)
        \STATE \textbf{Initialize Certain Sets:} Initialize $\sZ = \set{Z_{1},\cdots,Z_{n_{c}}}$ and numQueries via \algname{UoS-Explore} in Appendix \ref{app:uosExplore}. 
        \WHILE{numQueries $<$ maxQueries}
        \STATE \textbf{PCA on Each Cluster:} Solve $$\sS_k = \min_{U\in \bR^{D\times d}} \sum_{i : \hat{C}(x_i)=k} \|x_i-UU' x_i\|^2\;.$$
        \STATE \textbf{Obtain Test Point:} select $x_{T} \gets \argmin_{x \in \sX} \hat{\mu}(x)$
            \STATE \textbf{Assign $x_{T}$ to Certain Set:}
            \begin{adjustwidth}{1em}{0em} \vspace{-1em}
                \STATE Sort $\set{Z_{1},\cdots,Z_{n_{c}}}$ in order of most likely must-link (via subspace residual for $x_{T}$), query $x_{T}$ against representatives from $Z_{k}$ until must-link constraint is found or $k = n_{c}$. If no must-link constraint is found, set $\sZ \gets \set{Z_{1},\cdots,Z_{n_{c}},\set{x_{T}}}$ and increment $n_{c}$.
            \end{adjustwidth}
            \STATE \textbf{Impute Constraints:} Set $A_{ij} = A_{ji} = 1$ for $(x_{i},x_{j})$ in the same certain set and $A_{ij} = A_{ji} = 0$ for $(x_{i},x_{j})$ in different certain sets (do not impute for points absent from certain sets).
            \STATE \textbf{Estimate Labels:} $\hat{C} \gets$ \algname{SpectralClustering}($A$,$K$)
        \ENDWHILE
    \end{algorithmic}
\end{algorithm}

\begin{figure*}[h!]
    \centering
    \includegraphics[width=6.5in]{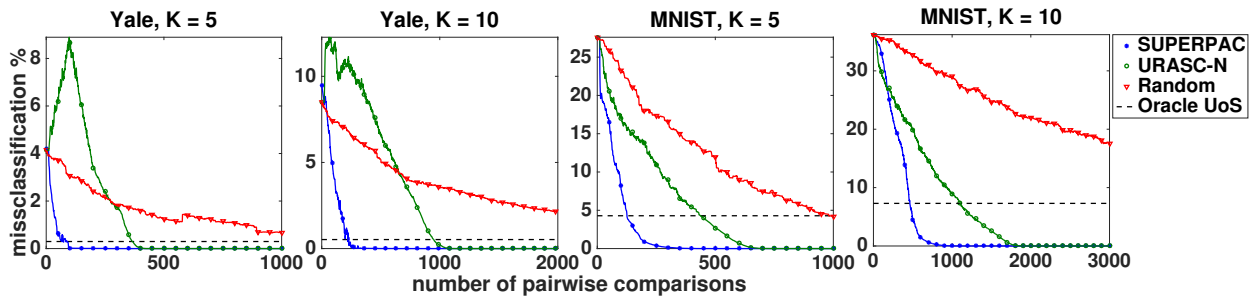}
    \vspace{-1em}
    \caption{Misclassification rate for Yale B and MNIST datasets with many pairwise comparisons. Left-to-right: 
    Yale B $K = 5$ (input from SSC), Yale B $K = 10$ (input from SSC), MNIST $K = 5$ (input from TSC), MNIST $K = 10$ (input from TSC).}
    \label{fig:yale_mnist_many}
\end{figure*}

\subsection{Pairwise Constrained Clustering with SUPERPAC}

We now describe SUPERPAC in more detail, our algorithm for PCC when data lie near a union of subspaces, given in Algorithm~\ref{alg:superpac}. The algorithm begins by initializing a set of disjoint certain sets, an optional process described in Appendix \ref{app:uosExplore}. 
Next our algorithm assigns the points most likely to be misclassified to certain sets by presenting a series of pairwise comparisons. Finally, we impute values onto the affinity matrix for all points in the certain sets and
perform spectral clustering. The process is then repeated until the maximum number of pairwise comparisons has been reached. 

Let $x_{T}$ be the test point chosen as the min-margin point.
Our goal is to assign $x_{T}$ to a certain set using as the fewest number of queries possible. For each certain set $Z_{k}$, the representative $x_{k}$ is chosen as the maximum-margin point within the set. Next, for each $k$, we let $U_{k}$ be the $d$-dimensional PCA estimate
of the matrix whose columns are the points $\set{x \in \sX: \hat{C}(x) = \hat{C}(x_{k})}$. We then query our test point $x_{T}$ against the representatives $x_{k}$ in order of residual $\norm{x_{T} - U_{k}U_{k}^{T}x_{T}}_{2}$ (smallest first). If a must-link constraint is found, we place $x_{T}$ in the corresponding certain set. Otherwise, we place $x_{T}$ in its own certain set and update the number of certain sets. Pseudocode for the complete algorithm is given in Algorithm \ref{alg:superpac}. As a technical note, we first normalize the input affinity matrix $A$ so that the maximum value is 2. For must-link constraints, we impute a value of 1 in the affinity matrix, while for cannot-link constraints we impute a 0. The approach of imputing values in the affinity matrix is
    common in the literature but does not strictly enforce the constraints. Further, we found in our experiments that imputing the maximum value in the affinity matrix resulted in unstable results. Thus, users must be careful to not only choose the correct constraints as noted in \cite{basu2004active}, but to incorporate these constraints in a way that allows for robust clustering. 

SUPERPAC can be thought of as an extension of ideas from PCC literature \cite{basu2004active,biswas2014active,xiong2016active} to leverage prior knowledge about the underlying geometry of the data. For datasets such as Yale B and MNIST, the strong subspace structure makes Euclidean distance a poor proxy for similarity between points in the same cluster, leading to the superior performance of our algorithm demonstrated in the following sections. This structure does not
exist in all datasets, in which case we do not expect our algorithm to outperform current PCC algorithms. The reader will note we made a choice to order the certain sets according to the UoS model; this is similar to the choice in \cite{xiong2016active} to query according to similarity, where our notion of similarity here is based on subspace distances. 
We found this resulted in significant performance benefits, matching our intuition that points are
clustered based on their nearest subspace. 
In contrast to \cite{biswas2014active,xiong2016active}, where the test point is chosen according to a global improvement metric, we choose
test points according to their classification margin. In our experiments, we found subspace margin to be a strong indicator of which points are misclassified, meaning that our algorithm rapidly corrects the errors that occur as a result of unsupervised subspace clustering. 

Finally, note that the use of certain sets relies on the assumption that the pairwise queries are answered correctly\textemdash an assumption that is common
in the literature \cite{basu2004active,mallapragada2008active,xiong2016active}. However, in \cite{xiong2016active}, the authors demonstrate that an algorithm based on certain sets still yields significant improvements under a small error rate. The study of robustly incorporating noisy pairwise comparisons is an interesting topic for further study.

\section{Experimental Results}
\label{sec:simulations}

We compare the performance of our method and the nonparametric version of the URASC algorithm (URASC-N)\footnote{In our experiments, the parametric version of URASC was found to be numerically unstable and did not have significantly different performance from URASC-N in the best cases.} over a variety of datasets. Note that while numerous PCC algorithms exist, URASC achieves both the best empirical results and computational complexity on a variety of datasets. We also compared with
the methods from \cite{basu2004active} and \cite{biswas2014active} but found both to perform significanly worse than URASC on all datasets considered, with a far greater computational cost in the case of \cite{biswas2014active}. We use a maximum query budget of $2K$ for \algname{UoS-Explore} and \algname{Explore}. For completeness, we also compare to random constraints, in which queries are chosen uniformly at random from the set of unqueried pairs. 

Finally, we compare against the oracle PCA classifier, which we now define. Let $U_{k}$ be the $d$-dimensional PCA estimate of the points whose true label $C(x) = k$. Then the oracle label is
$\hat{C}_{o}(x) = \argmin_{k \in [K]} \norm{x - U_{k}U_{k}^{T}x}_{2}$.
This allows us to quantitatively capture the idea that, because the true classes are not perfectly low-rank, some points would not be clustered with the low-rank approximation of their own true cluster. In our experiments, we also compared with oracle robust PCA \cite{candes2011robust} implemented via the augmented Lagrange multiplier method \cite{lin2011linearized} but did not find any improvement in classification error. 

\begin{figure}[t]
    \centering
    \includegraphics[width=0.7\columnwidth]{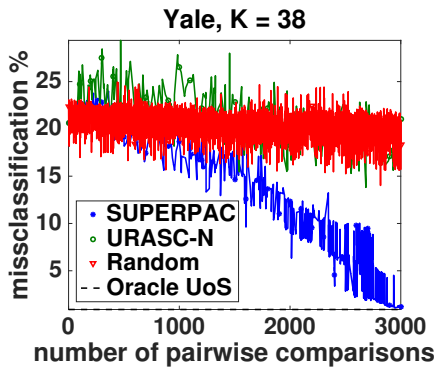}
    \caption{Misclassification rate versus number of pairwise comparisons for extended Yale face database B with $K = 38$ subjects. Input affinity matrix is taken from SSC-OMP.}
    \label{fig:yaleFull}
    \vspace{-1em}
\end{figure}

\paragraph{Datasets}

We consider five datasets commonly used as benchmarks in
the subspace clustering literature\footnote{The validity of the UoS assumption for two of these datasets is investigated in \cite{elhamifar2013sparse,heckel2015robust}.}, with a summary of the datasets and their relevant parameters are given in Table \ref{tab:datasets}. 
The Yale B dataset consists of 64 images of size 192 $\times$ 168 of each of 38 different subjects under a variety of lighting conditions. For values of $K$ less than 38, we follow the methodology of \cite{zhang2012hybrid} and perform clustering on 100 randomly selected subsets of size $K$. We choose $d=9$ as is common in the literature
\cite{elhamifar2013sparse,heckel2015robust}. 
The MNIST handwritten digit database test dataset consists of 10,000 centered 28 $\times$ 28 pixel images of handwritten digits 0-9. We follow a similar methodology to the previous section and select 100 random subsets of size $K$, using subspace dimension $d=3$ as in \cite{heckel2015robust}. 
The COIL-20 dataset \cite{nene1996coil20} consists of 72 images of size 32 $\times$ 32 of each of 20 objects. The COIL-100 dataset \cite{nene1996coil100} contains 100 objects (distinct from the COIL-20 objects) of the same size and with the same number of images of each object. For both datasets, we use subspace dimension $d = 9$.
Finally, we apply our algorithm to the USPS dataset provided by \cite{cai11graph}, which contains 9,298 \textit{total} images of handwritten digits 0-9 of size 16 $\times$ 16 with roughly even label distribution. We again use subspace dimension $d = 9$. 

\paragraph{Input Subspace Clustering Algorithms}

A major strength of our algorithm is that it is agnostic to the initial subspace clustering algorithm used to generate the input affinity matrix. To demonstrate this fact, we apply our algorithm with an input affinity matrix obtained from a variety of subspace clustering methods, summarized in Table \ref{tab:datasets}. Note that some recent algorithms are not included in the simulations here. However, the simulations show that our algorithm works well with \textit{any} initial
clustering, and hence we expect similar results as new algorithms are developed.

\begin{figure*}[h!]
    \hspace{1in}
    \includegraphics[width=6.5in]{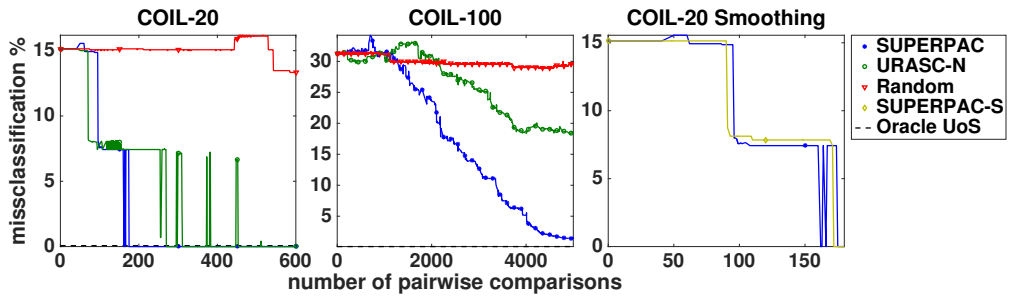}
    \vspace{-2em}
    \caption{Misclassification rate versus number of pairwise comparisons for COIL-20 ($K = 20$) and COIL-100 ($K = 100)$) databases. Input affinity matrix is taken from EnSC. Rightmost plot shows proposed smoothing heuristic.}
    \label{fig:coil}
\end{figure*}

\paragraph{Experimental Results}

\begin{table}[t]
    \centering
    \begin{tabular}{ | c || c | c | c | c | c | c | c | c | }
        \hline 
        Dataset & $N$ & $K$ & $D$ & $d$ \\
        \hline
        Yale & 320-2432 & 5,10,38 & 2016 & 9 \\
        \hline
        MNIST & 500-1000 & 5,10 & 784 & 3 \\
        \hline
        COIL-20 & 1440 & 20 & 1024 & 9 \\
        \hline
        COIL-100 & 7200 & 100 & 1024 & 9 \\
        \hline
        USPS & 9298 & 10 & 256 & 15 \\
        \hline
    \end{tabular}
    \caption{Datasets used for experiments with relevant parameters; $N$: total number of samples, $K$: number of clusters, $D$: ambient dimension, $d$: estimated subspace dimension.}
    \label{tab:datasets}
\end{table}

Fig. \ref{fig:yale_mnist_many} shows the clustering error versus the number of pairwise comparisons for the Yale and MNIST datasets. The input affinity matrix is obtained by running SSC for the Yale datset and by running TSC for the MNIST dataset.
The figure clearly demonstrates the benefits of leveraging UoS structure in constrained clustering\textemdash in all cases, SUPERPAC requires roughly \textit{half} the number of queries needed by URASC to achieve perfect clustering. For the Yale dataset with $K = 5$, roughly $2Kd$ queries are required to surpass oracle performance, and
for $K = 10$ roughly $3Kd$ queries are required. Note that for the Yale dataset, the clustering error \textit{increases} using URASC. This is due to the previously mentioned fact that imputing the wrong constraints can lead to worse clustering
performance. For sufficiently many queries, the error decreases as expected. 
Fig. \ref{fig:yaleFull} shows the misclassification rate versus number of points for all $K = 38$ subjects of the Yale databse, with the input affinity matrix taken from SSC-OMP \cite{you2016scalable}. We space out the markers for clearer plots. In this case, URASC performs roughly the same as random query selection, while SUPERPAC performs significantly better.

\begin{figure}[t]
    \vspace{-2mm}
    \centering
    \includegraphics[width=0.7\columnwidth]{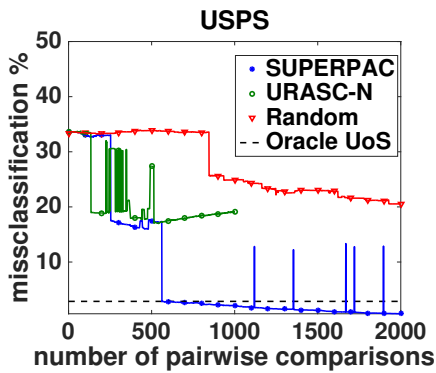}
    \caption{Misclassification rate versus number of pairwise comparisons for USPS dataset with $K = 10$ digits, 9,298 total samples. Input affinity matrix is taken from EnSC. URASC did not complete after 48 hours of run time.}
    \label{fig:usps}
    \vspace{-2mm}
\end{figure}

\begin{figure}[t]
    \vspace{-2mm}
    \centering
    \includegraphics[width=0.7\columnwidth]{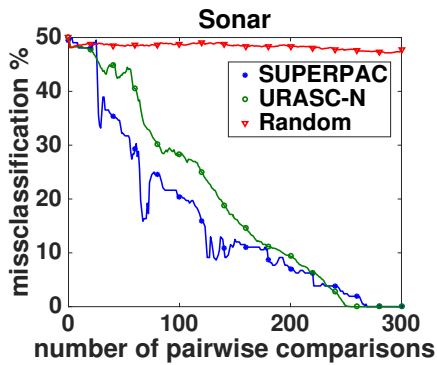}
    \caption{Misclassification rate for Sonar dataset from \cite{xiong2016active}, where there is not reason to believe the clusters have subspace structure. We are still very competitive with state-of-the-art.}
    \label{fig:sonar}
    \vspace{-2mm}
\end{figure}

Fig. \ref{fig:coil} demonstrates the continued superiority of our algorithm in the case where UoS structure exists. In the case of COIL-20, the clustering is sometimes unstable, alternating between roughly 0\% and 7\% clustering error for both active algorithms. This further demonstrates the observed phenomenon that spectral clustering is sensitive to small perturbations. To avoid this issue, we kept track of the $K$-subspaces cost function (see \cite{bradley2000kplane}) and ensured the cost decreased at every iteration. We refer to this added heuristic as SUPERPAC-S in the figure. The incorporation of this heuristic into our algorithm is a topic for further study.


Fig. \ref{fig:usps} shows the resulting error on the USPS dataset, again indicating the superiority of our method. Note that $N$ is large for this dataset, making spectral clustering computationally burdensome. Further, the computational complexity of URASC is
dependent on $N$. As a result, URASC did not complete 2000 queries in 48 hours of run time when using 10 cores, so we compare to the result after completing only 1000 queries. Finally, in Fig. \ref{fig:sonar}, we demonstrate that even on data without natural subspace structure, SUPERPAC performs competitively with URASC. 

\section{Conclusion}

We have presented a method of selecting and incorporating pairwise constraints into subspace clustering that considers the underlying geometric structure of the problem. The union of subspaces model is often used in computer vision applications where it is possible to request input from human labelers in the form of pairwise constraints. We showed that labeling is often necessary for subspace classifiers to achieve a clustering error near zero; additionally, these constraints can be chosen intelligently to improve
the clustering procedure overall and allow for perfect clustering with a modest number of requests for human input. 


Developing techniques for handling noisy query responses will allow extension to undersampled or compressed data. One may assume that compressed data would be harder to distinguish, leading to noisier query responses.  
Finally, we saw that for datasets with different types of cluster structure, the structure assumptions of each algorithm had direct impact on performance; in the future we plan to additionally develop techniques for learning from unlabeled data whether the union of subspace model or a standard clustering approach is more appropriate.

\section*{\Large Appendices}
\appendix
\section{Proof of Theorem \ref{thm:jl}}
\label{app:thm}

The proof relies on theorem 5.2.1 from \cite{vershynin2016course}, restated below.
\begin{thm}{(Concentration on Gauss space)}
    Consider a random vector $X \sim \sN(0,\sigma^{2}I_{D})$ and a Lipschitz function $f: \bR^{D} \rightarrow \bR$. Then for every $t \geq 0$,
    \begin{equation*}
        \bP\set{\left| f(X) - \bE f(X) \right| \geq t} \leq 2 \exp\left( -\frac{ct^{2}}{\sigma^{2}\norm{f}_{\mathrm{Lip}}^{2}} \right),
    \end{equation*}
    where $\norm{f}_{\mathrm{Lip}}$ is the Lipschitz constant of $f$.
    \label{thm:gaussConc}
\end{thm}
First consider the numerator and note that $y - P_{1}y = P_{1}^{\perp}y \sim \sN(0,\sigma^{2}P_{1}^{\perp})$ with
\begin{equation*}
    \bE \norm{P_{1}^{\perp}y}^{2} = \sigma^{2}(D-d).
\end{equation*}
Let $f(z) = \norm{Pz}_{2}$, where $P$ is an arbitrary projection matrix. In this case, $\norm{f}_{\mathrm{Lip}} = 1$, as $f$ is a composition of 1-Lipschitz functions, which is also 1-Lipschitz. Further, by Exercise 5.2.5 of \cite{vershynin2016course}, we can replace $\bE \norm{X}_{2}$ by $\left( \bE \norm{X}_{2}^{2} \right)^{1/2}$ in the concentration inequality. Applying Thm. \ref{thm:gaussConc} to the above, we see that
\begin{equation}
    \bP\set{\left| \norm{P_{1}^{\perp}y} - \sqrt{\sigma^{2}(D-d)} \right| \geq t} \leq 2 \exp\left( -\frac{ct^{2}}{\sigma^{2}} \right).
    \label{eq:num}
\end{equation}
Similarly, for the denominator, note that $y - P_{2}y = P_{2}^{\perp}y \sim \sN(P_{2}^{\perp}x,\sigma^{2}P_{2}^{\perp})$ with
\begin{equation*}
    \bE \norm{P_{2}^{\perp}y}^{2} = \sigma^{2}(D-d) + \gamma^{2}.
\end{equation*}
Since $P_{2}^{\perp}y$ is no longer centered, we let $g(z) = z + P_{2}^{\perp}x$, which also has $\norm{g}_{\mathrm{Lip}} = 1$. Applying Thm. \ref{thm:gaussConc} to the centered random vector $\bar{y} \sim \sN(0,\sigma^{2}P_{2}^{\perp})$ with Lipschitz function $h = f \circ g$, we have that
\begin{equation}
    \bP\set{\left| \norm{P_{2}^{\perp}y} - \sqrt{\sigma^{2}(D-d) + \gamma^{2}} \right| \geq t} \leq 2 \exp\left( -\frac{ct^{2}}{\sigma^{2}} \right).
    \label{eq:denom}
\end{equation}
Letting $t = \varepsilon \sqrt{\sigma^{2}(D-d)}$ in \eqref{eq:num} and $t = \varepsilon \sqrt{\sigma^{2}(D-d) + \gamma^{2}}$ in \eqref{eq:denom} yields
\begin{equation*}
    (1-\varepsilon) \sqrt{\sigma^{2}(D-d)} \leq \norm{P_{1}^{\perp}y} \leq (1+\varepsilon) \sqrt{\sigma^{2}(D-d)}
\end{equation*}
and
\begin{eqnarray*}
    (1-\varepsilon) \sqrt{\sigma^{2}(D-d) + \gamma^{2}} \leq \norm{P_{2}^{\perp}y} \\
    \leq (1+\varepsilon) \sqrt{\sigma^{2}(D-d) + \gamma^{2}},
\end{eqnarray*}
each with probability at least $1 - 2 \exp\left( -c\varepsilon^{2}(D-d) \right)$ (since $\gamma > 0$). Applying the union bound gives the statement of the theorem.

\section{Proof of Corollary \ref{cor:mu}}
\label{app:cor}

    We have from Thm. \ref{thm:jl} that $$\mu(y_2) \leq \frac{(1+\varepsilon)\sqrt{\sigma^{2}(D-d)}}{(1-\varepsilon)\sqrt{\sigma^{2}(D-d) + \gamma_2^{2}}}$$ and $$\frac{(1-\varepsilon)\sqrt{\sigma^{2}(D-d)}}{(1+\varepsilon) \sqrt{\sigma^{2}(D-d) + \gamma_1^{2}}} \leq \mu(y_1)$$ with probability at least $1 - 4e^{-c\varepsilon^{2}(D-d)}$. Therefore if we get the upper bound of $\mu(y_2)$ to be smaller than the lower bound of $\mu(y_1)$, we are done. Rearranging this desired inequality we see that we need \begin{equation}
    \gamma_{1}^{2} < \beta^{4}\gamma_{2}^{2} - (1-\beta^{4}) \sigma^{2}(D-d).
\end{equation} where $\beta = (1-\varepsilon)/(1+\varepsilon)$. Let $\varepsilon$ be such that $\beta^4 = 5/6$, and let $\gamma_1^2 =  \sin^2(\phi_1) + \delta s$ as in the theorem. Then we wish to select $\delta$ to satisfy 
 \begin{equation}
    \delta  < \frac{\frac{5}{6} \gamma_{2}^{2} - \sin^2(\phi_1) - \frac{1}{6} \sigma^{2}(D-d)}{s}.
\end{equation} Applying concentration with $\gamma_2^2$, we have that $\gamma_2^2 \geq (1-\xi)^2 s$ with probability at least $1-e^{-c \xi^2 d s}$ where $c$ is an absolute constant. Therefore taking $\xi$ to be such that $(1-\xi)^2=6/7$, we require $$\delta < \frac{\frac{5}{7} s - \sin^2(\phi_1) - \frac{1}{6} \sigma^{2}(D-d)}{s} = \frac{5}{7} - \frac{1}{\tau}$$ where we used the definition of $\tau$ in the theorem. To quantify the probability we need the appropriate values for $\varepsilon$ and $\xi$; we lower bound both with simple fractions: $1/50 < \varepsilon$ where $\left((1-\varepsilon)/(1+\varepsilon)\right)^4 = \beta = 5/6$ and $7/100 < \xi$ where $(1-\xi)^2 = 6/7$. Applying the union bound with the chosen concentration values implies that $\mu(y_1)>\mu(y_2)$ holds with probability at least $1-e^{-c \left(\frac{7}{100}\right)^2 d s} - 4e^{-c\left(\frac{1}{50}\right)^{2}(D-d)}$.

\section{\algname{UoS Explore} Algorithm}
\label{app:uosExplore}

In this section, we describe the process of initializing the certain sets. Note that this step is not necessary, as we could initialize all certain sets to be empty, but we found it led to improved performance experimentally. A main distinction between subspace clustering and the general clustering problem is that in the UoS model points can lie arbitrarily far from each other but still be on or near the same subspace. For this reason, the \algname{Explore} algorithm from \cite{basu2004active} is unlikely to quickly find points from different clusters in an efficient manner. Here we define an analogous algorithm for the UoS case, termed \algname{UoS-Explore}, with pseudocode given in Algorithm
\ref{alg:UoS-Explore}. The goal of \algname{UoS-Explore} is to find $K$ certain sets, each containing as few points as possible (ideally a single point), allowing us to more rapidly assign test points to certain sets in the SUPERPAC algorithm. We begin by selecting our test point $x_{T}$ as the most certain point, or the point of \textit{maximum} margin and placing it in its own certain set. We then iteratively select $x_{T}$ as the point of maximum margin that (1) is not in any certain set and (2) has a
different cluster estimate from all points in the certain sets. 
\begin{algorithm}[t]
    \caption{\algname{UoS-Explore}}
    \label{alg:UoS-Explore}
    \begin{algorithmic}
        \STATE \textbf{Input:} $\sX = \set{x_{1},x_{2},\dots,x_{N}}$: data, $K$: number of subspaces, $d$: dimension of subspaces, $A$: affinity matrix, maxQueries: maximum number of pairwise comparisons
        \STATE \textbf{Estimate Labels:} $\hat{C} \gets$ \algname{SpectralClustering}($A$,$K$)
        \STATE \textbf{Calculate Margin:} Calculate margin and set \\
        \hspace{1em} $x_{\vee} \gets \argmax_{x \in \sX} \hat{\mu}(x)$ (most confident point)
        \STATE \textbf{Initialize Certain Sets:} $Z_{1} \gets x_{\vee}$, $\sZ \gets \set{Z_{1}}$, \\
        \hspace{1em} numQueries $\gets$ 0, $n_{c} \gets 1$
        \WHILE{$n_{c} < K$ \AND numQueries $<$ maxQueries}
        \STATE \textbf{Obtain Test Point:} Choose $x_{T}$ as point of maximum margin such that $\hat{C}(x_{T}) \neq \hat{C}(x \in Z_{k})$ for any $k$. If no such $x_{T}$ exists, choose $x_{T}$ at random.
        \STATE \textbf{Assign $x_{T}$ to Certain Set:}
        \begin{adjustwidth}{1em}{0em} \vspace{-1em}
            \STATE Sort $\set{Z_{1},\cdots,Z_{n_{c}}}$ in order of most likely must-link (via subspace residual for $x_{T}$), query $x_{T}$ against representatives from $Z_{k}$ until must-link constraint is found or $k = n_{c}$. If no must-link constraint found, set $\sZ \gets \set{Z_{1},\cdots,Z_{n_{c}},\set{x_{T}}}$ and increment $n_{c}$.
            \end{adjustwidth}
        \ENDWHILE
    \end{algorithmic}
\end{algorithm}
If no such point exists, we choose uniformly at random from all points not in any certain set. This point is queried against a single representative from each certain set according to the UoS model as above until either a must-link is found or all set representatives have been queried, in which case $x_{T}$ is added to a new certain set. This process is repeated until either $K$ certain sets have been created or a terminal number of queries have been used. As points of maximum margin are more likely to be correctly clustered than other points in the set, we expect that by choosing points whose estimated labels indicate they do not belong to any current certain set, we will quickly find a point with no must-link constraints. In our simulations, we found that this algorithm finds at least one point from each cluster in nearly the lower limit of $K(K-1)/2$ queries on the Yale dataset.

\section*{Acknowledgements}

This work was supported by NSF F031543-071159-GRFP and US ARO Grant W911NF1410634.

\bibliography{Bibliography}
\bibliographystyle{icml2017}

\end{document}